\documentclass[titlepage,oneside,12pt]{article}

\usepackage{multirow}
\usepackage{gensymb}

\usepackage{titling}

\usepackage[tiny,rm]{titlesec}

\newpagestyle{trbstyle}{
\sethead{}{}{\thepage}
}
\newcommand{\comment}[1]{}

\newlength\mylen

\usepackage{url}
\usepackage{hyperref}
\hypersetup{
  colorlinks   = true,
  linkcolor    = black,
  citecolor   = black
}

\usepackage{enumitem}
\setlist{nolistsep}

\titleformat{\section}{\bfseries}{}{0pt}{\uppercase}
\titlespacing*{\section}{0pt}{12pt}{*0}
\titleformat{\subsection}{\bfseries}{}{0pt}{}
\titlespacing*{\subsection}{0pt}{12pt}{*0}
\titleformat{\subsubsection}{\itshape}{}{0pt}{}
\titlespacing*{\subsubsection}{0pt}{12pt}{*0}

\usepackage{enumitem}
\setlist[1]{labelindent=0.5in,leftmargin=*}
\setlist[2]{labelindent=0in,leftmargin=*}

\usepackage{ccaption}
\usepackage{amsmath}
\makeatletter
\renewcommand{\fnum@figure}{\textbf{FIGURE~\thefigure} }
\renewcommand{\fnum@table}{\textbf{TABLE~\thetable} }
\makeatother
\captiontitlefont{\bfseries \boldmath}
\captiondelim{\;}

\usepackage{times}
\usepackage{float}
\usepackage{lineno}

\usepackage[T1]{fontenc}
\usepackage{textcomp}
\usepackage{caption} 
\usepackage[font=bf]{caption}

\usepackage[sort,numbers]{natbib}

\setcitestyle{round}

\usepackage{graphicx}

\begin{document}

\title{\large\textbf{REAL-WORLD MAPPING OF GAZE FIXATIONS USING INSTANCE SEGMENTATION FOR ROAD CONSTRUCTION SAFETY APPLICATIONS}}
\author{
Idris Jeelani \\ Department of Civil, Construction, and Environmental Engineering\\ North Carolina State University\\ 2501 Stinson Drive, Raleigh, NC 27695, USA \\ Tel:919-619-5926; Email:idrisj@ncsu.edu \\
\\Khashayar Asadi (Corresponding author)\\Department of Civil, Construction, and Environmental Engineering\\ North Carolina State University\\ 2501 Stinson Drive, Raleigh, NC 27695, USA \\ Tel:919-917-0326; Email:kasadib@ncsu.edu \\
\\Hariharan Ramshankar\\ Security Sytem Engineer\\ M.C. Dean\\ 3066 Scott Blvd, Santa Clara CA 95054\\ Tel:919-633-8439; Email:hariharan.ramshankar@gmail.com \\
\\
Kevin Han\\
Department of Civil, Construction, and Environmental Engineering\\ North Carolina State University\\ 2501 Stinson Drive, Raleigh, NC 27695, USA \\ 
Tel:510-229-9648; Email:kevin\_han@ncsu.edu \\
\\
Alex Albert\\
Department of Civil, Construction, and Environmental Engineering\\ North Carolina State University\\ 2501 Stinson Drive, Raleigh, NC 27695, USA \\ 
Tel:919-515-7208; Email:alex\_albert@ncsu.edu \\
\\
6105 words + 3 tables x 250 words = 6855}
\thispagestyle{empty}
\maketitle

\clearpage
\section{Abstract}
Research studies have shown that a large proportion of hazards remain unrecognized, which expose construction workers to unanticipated safety risks. Recent studies have also found that a strong correlation exists between viewing patterns of workers, captured using eye-tracking devices, and their hazard recognition performance. Therefore, it is important to analyze the viewing patterns of workers to gain a better understanding of their hazard recognition performance. This paper proposes a method that can automatically map the gaze fixations collected using a wearable eye-tracker to the predefined areas of interests. The proposed method detects these areas or objects (i.e., hazards) of interests through a computer vision-based segmentation technique and transfer learning. The mapped fixation data is then used to analyze the viewing behaviors of workers and compute their attention distribution. The proposed method is implemented on an under construction road as a case study to evaluate the performance of the proposed method.

{\bf Keywords:} Hazard recognition, road construction safety, transfer learning, eye-tracking, machine vision
\clearpage
\section{INTRODUCTION}

With an average of nine fatalities every day, construction is one of the most dangerous industries for which to work \cite{albert2014experimental}.  In the United States, the construction industry accounts for approximately 16\% of all occupational fatalities \cite{bureau2013case}. Worldwide, more than 60,000 workers fatalities are reported from construction sites every year \cite{lingard2013occupational}. Not surprisingly, the Center for Construction Research and Training (CPWR) estimates that a worker is 75\% likely to suffer a disabling injury during a 45-year career in the construction industry \cite{schwatka2016safety}. Despite significant advancements in safety enforcement, training, and monitoring, desirable levels of safety performance are not attained. 

To prevent safety incidents, current practices largely rely on the ability of workers and professionals to identify and manage safety hazards. Unfortunately, research studies have demonstrated that construction workers and professionals fail to recognize hazards at an unacceptable rate \cite{jeelani2017development,bahn2013workplace,carter2006safety,jeelani2016development}. The poor recognition of hazards can result in unanticipated hazard exposure and can substantially increase the likelihood of accidents and injuries. To address this safety issue, past studies have focused on examining several factors including training, education, and management support that may indirectly influence hazard identification and management performance \cite{abdelhamid2000identifying,mitropoulos2005systems,rajendran2009impact,balali2014improved}.  However, proximal factors associated with poor hazard identification and management at the work interface has received very little attention. There is a lack of knowledge regarding factors that influence hazard identification when workers examine the workplace during safety planning session.

Jeelani et al. and Dzeng et al. \cite{jeelani2018automating,dzeng2016using, doi:10.1061/(ASCE)CO.1943-7862.0001589} reported that viewing patterns of workers are correlated with their hazard recognition performance. Asadi et al. \cite{asadi2017advancing} also found that the amount of time participants spent looking at hazard was closely related to the number of hazards identified correctly. Hence, there is value in studying and analyzing the viewing patterns of workers to understand the parameters of a good hazard search.

Although the above studies paved the way to study the visual search behavior of construction workers and the resulting impact on their ability to recognize hazards, all of these studies recorded the viewing behavior of workers while they searched for hazards in still photographs. However, viewing behavior and attentional distribution of workers might be significantly different when they are in a real construction environment as compared to their attentional distribution while viewing a still photograph. Studying visual search patterns of workers while viewing a real environment is expected to provide a better understanding of the attentional distribution of workers and its impact on hazard recognition performance \cite{jeelani2018scaling}. This data, when collected on a large scale, can be used to provide effective training solutions and design effective hazard recognition and management practices. Moreover, recent advances in computer vision techniques \cite{Girshick_2014_CVPR,girshick2015fast,long2015fully,lightdeep} that detect and segment objects can automate the analysis of visual search patterns with respect to hazards. 

\begin{figure}[H]
\begin{center}\includegraphics[width=7in]{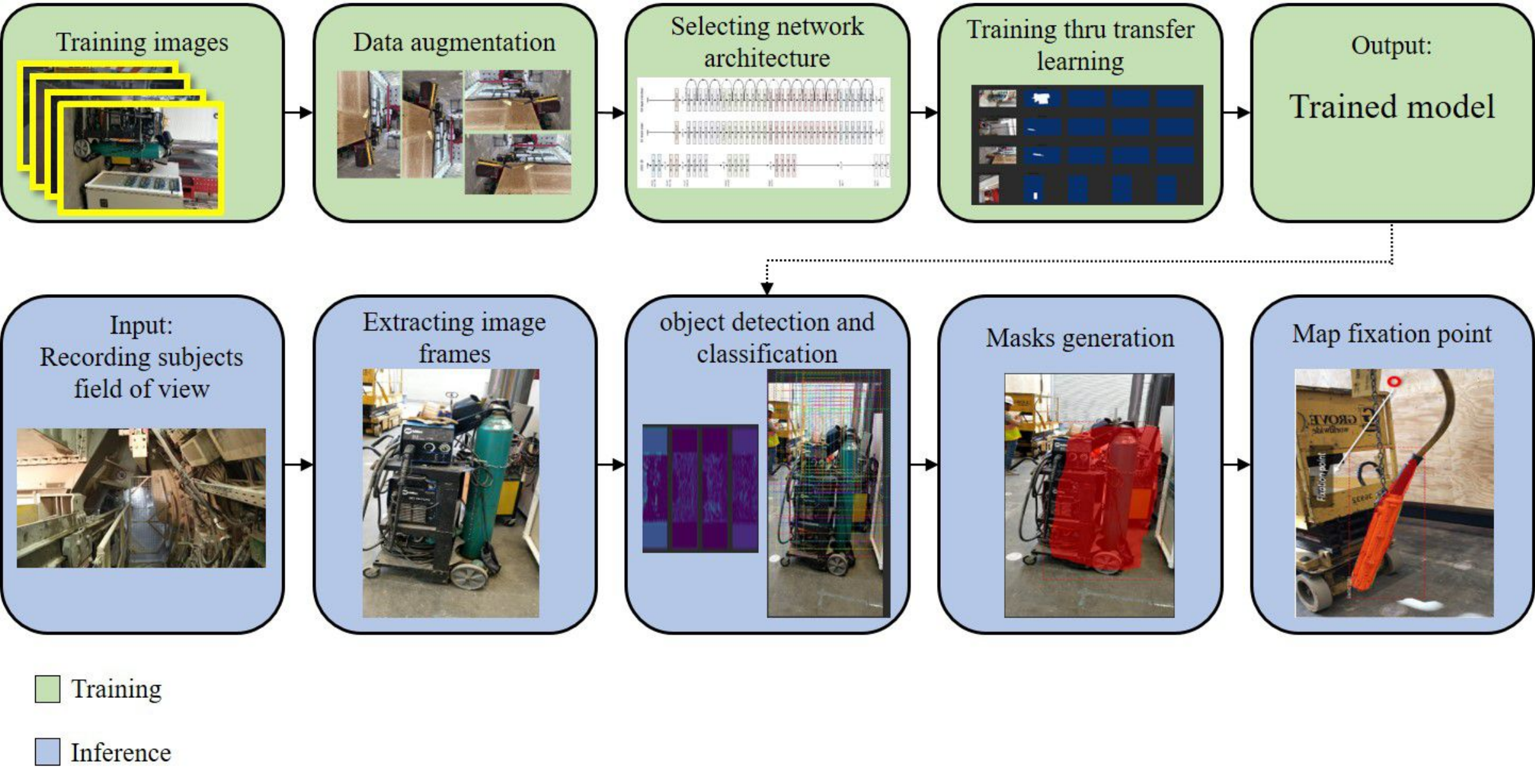}\end{center}
\caption{Method overview}
\label{fig:0}
\end{figure}

This study proposes the development of a system for automatically mapping the gaze fixations from eye-tracking glasses to the predefined areas of interests. The proposed method uses a computer-vision based object segmentation technique, recurrent convolution neural network (RCNN) \cite{DBLP:journals/corr/HeGDG17}, and transfer learning to detect predefined objects (i.e., hazards). Workers wear eye-tracking glasses that capture videos and eye movement data while they move in a road construction site. This video is passed to the trained model for inferencing and masking objects of interest (i.e., hazards) when a predefined area of interest is detected in the test video using an instance segmentation framework. Using the fixation data from the eye-tracking glasses, each worker's viewing behavior (i.e., attention distribution, fixation points, etc.) is monitored and analyzed automatically. The system can also be used to predict the detectability of various hazards in a road construction work environment. Figure \ref{fig:0} shows a brief overview of the system.

\section{BACKGROUND}
\subsection{Visual search and eye-tracking }
Visual search is a task of scanning an environment to search for a specific object (or type of objects that are contextually related) among other objects known as distractors \cite{jeelani2018automating}. For example, a shopper looking for their proffered cereal in a supermarket aisle or a doctor scanning through an MRI scan to look for tumors. Similarly, a construction professional looking for potential hazards in their work zone is also a visual search task. Several factors like the similarity in distractors and targets \cite{guest2011time}, age differences \cite{madden2004age,humphrey1997age}, professional experience, and the type of search being conducted, significantly affect the results of a visual search.   

Eye tracking is one of the effective tools for monitoring and analyzing a subject's visual search.  Cameras and infrared sensors have been used in different fields to record and monitor eye movements that are subsequently used to compute the attentional distribution and viewing patterns of human subjects. This helps to objectively measure the amount of attention received by different stimuli \cite{salamati2012simulator,fisher2007empirical}.Eye-tracking technology has found applications in medicine, aviation, education, and transportation \cite{steelman2011modeling,tien2014eye,brimley2014use}. However, eye tracking studies are still in the nascent stages of construction. Nonetheless, eye tracking studies in real-world environments can help us to analyze visual search processes, such as hazard recognition \cite{li2018driver}, that depends on hidden mental process and stores tacit schemas which vary across individuals and are largely inaccessible to researchers \cite{bojko2013eye}.

\subsection{Vision-based object recognition}
In construction, computer vision has drawn attention because of its ability to support automation efforts and the continuous monitoring of construction operations \cite{computing,ASADI2018470,han2017potential,han2015formalized,asadi2018real,asadi2017perspective}. Object recognition has been and still is, one of the major topics of interest in the computer vision community. Object recognition can be performed through object classification or scene segmentation. While semantic segmentation is computationally more intensive for real-time applications, it would result in more accurate boundaries of the objects in the scene \cite{asadi2018building}. For instance, object detection would classify all objects in the scene using boxes whereas semantic segmentation will divide the scene by putting boundaries around the objects.

Even though Convolutional Neural Networks (CNN) had been around for years \cite{lecun1998gradient}, it has not been used for bounding-box object classification until recently. Girshick et al. \cite{Girshick_2014_CVPR} \cite{girshick2015fast} combined image region proposal scheme with CNN as a classifier and performed near real-time inference. The R-CNN limitations such as slow test time and training pipeline complexity were resolved in the later versions \cite{girshick2015fast,ren2015faster}. In fast R-CNN object detection method, the computation of convolutional layers was shared between region proposals of an image which made the inference time 25 times faster \cite{girshick2015fast}. Faster R-CNN\cite{ren2015faster} improved computational speed up to 250x by inserting a region proposal network (RPN) after the last convolutional layer (no external region proposal was needed).

CNN-based frameworks for semantic segmentation have also achieved near real-time inference time although the common drawback still is the computational load. On the other hand, Long et al. \cite{long2015fully} proposed a semantic segmentation method that takes exiting deep CNNs \cite{szegedy2015going,krizhevsky2012imagenet,simonyan2014very} into a fully convolutional network (FCN) and performed segmentation from their learning representations. Badrinarayanan et al. \cite{badrinarayanan2015segnet} proposed a semantic segmentation method that is based on a very large encoder-decoder model, performing pixel-wise labeling and resulting in a very large number of computations.

One of the main goals of this study is the real-time performance of the proposed system. Computationally lighter convolutional networks have been presented in \cite{howard2017mobilenets,paszke2016enet}. He et al. \cite{DBLP:journals/corr/HeGDG17} presented Mask RCNN  approach that builds on Faster R-CNN with a parallel mask generation network on detected Regions of Interest(ROI). Mask RCNN adds very little computational overhead, running at 5 fps, and is intuitive in nature. The results were presented on COCO dataset \cite{lin2014microsoft} by combining the best of object detection with semantic segmentation to obtain instance segmentation which is relevant for this study. 

\section{OBJECTIVE AND POINT OF DEPARTURE}
\label{objective}
\noindent The objective of this study is to develop and test an automated system that maps the gaze fixations of workers on a construction site and analyze it to determine their visual attention distribution. The key steps are:
\begin{enumerate}
	\item Predefining hazards (AOIs)
    \item Training a model to detect AOI in test videos using transfer learning
    \item Mapping fixations with respect to detected AOIs
	\item Analyzing the gaze behaviors to determine attention distribution of workers, with respect to predefined hazards
    \item Computing various visual search metrics that define individual viewing patterns, to be used for further analysis or feedback generation.
\end{enumerate}

This study represents the first effort to combine eye-tracking technology and machine learning techniques to develop and test a prototype system that automates eye-tracking data analysis of real-world fixations. The study seeks to advance theoretical and practical knowledge related to improving hazard recognition levels within the construction and the use of computer vision techniques and eye-tracking technology to automate the analysis of eye tracking in real environments.

\section{SYSTEM DEVELOPMENT}
\label{System_Description}
The proposed system is developed in five stages as detailed below.

\subsection{Selecting the framework}
When a worker scans a scene, there are a vast variety of areas over which their gaze passes. The authors are interested in obtaining metrics whenever the worker's gaze is fixated on a predefined hazard. In simple words, this study is interested in knowing if and for how long a person is looking at a particular area of interest (AOI).

\begin{figure}[H]
\begin{center}\includegraphics[width=1.6in]{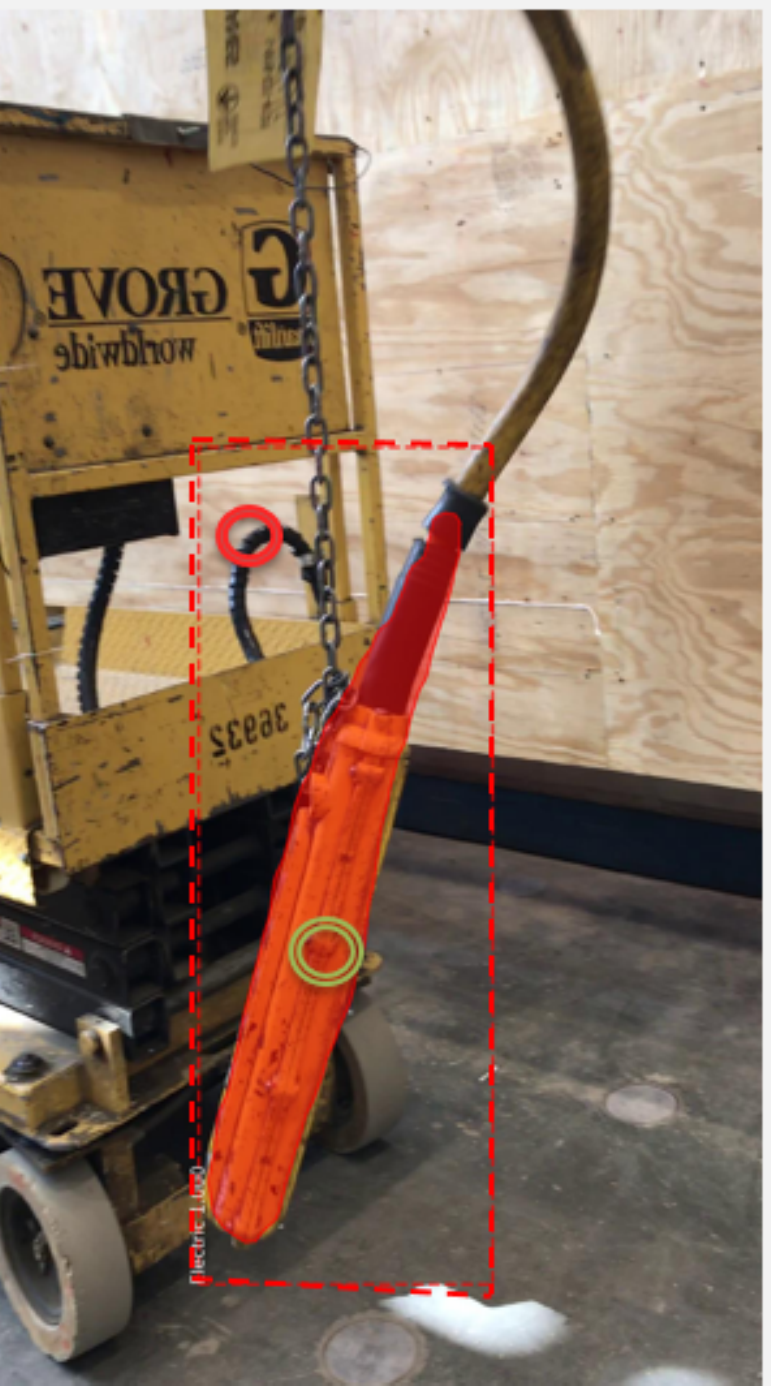}\end{center}
\caption{Instance segmentation provides better accuracy for eye-tracking analysis}
\label{fig_1}
\end{figure}

To this end, there is a need for accurate and real-time detection of objects (or AOIs) in the worker's view (i.e., First person view or FPV) that indicates whether or not an object is present in the scene. Then it is evaluated whether the gaze position is on that object (or within the AOI) (i.e is the subject looking at the object), if so, then the relevant eye tracking metrics discussed are calculated later. The first task is best done using an object detection framework and the second part is well suited for semantic segmentation. For example, in Figure \ref{fig_1}, an object detection framework would give a bounding box around the object of interest (an electric hazard in this case), and both green and red fixation points would be detected as being on the object. However, in reality, only the green fixation point is on the object (i.e a person is looking at the object). Whereas the red fixation point indicates that the person is looking somewhere else, yet both are within the bounding box. Therefore, the ideal framework would combine the benefits of object detection and segmentation to result in instance segmentation, which is why Mask R-CNN \cite{DBLP:journals/corr/HeGDG17} is preferred as the core framework in this study.\\

\subsection{Data collection \& preparation}
\subsubsection{Dataset description}
A new dataset is constructed for the proposed method. The main objective of the instance segmentation task is to detect and segment predetermined objects. For demonstration and validation, three classes of objects are chosen including a motion hazard (denoted as H1. e.g., excavator), electrical hazard (denoted as H2. e.g., cables), and mechanical/electrical hazard (denoted as H3. e.g., generator). The data is collected from a live construction site in Raleigh, North Carolina. 1,000 image frames are collected and annotated at a resolution of $1920\times1080$, later resized to $1024\times1024$ pixels with a square aspect ratio during training. 

\subsubsection{Annotation}
The data are annotated manually for each of the three constituent classes of chemical, electrical, and mechanical hazards. A GUI-based tool is used to ease the process.  Deep learning models need to be trained on a large dataset. A small data set could cause over-fitting (i.e., performing well on the training data but not on the testing data). To deal with this challenge of having a limited number of the training images, label-preserving data augmentation of flipping and rotation are used. These two methods have also been applied simultaneously to generate more samples. Each image in the original dataset could generate four new images through this data augmentation approach.

\subsection{Transfer learning and network training}
\begin{figure}[H]
\begin{center}\includegraphics[width=2.7in]{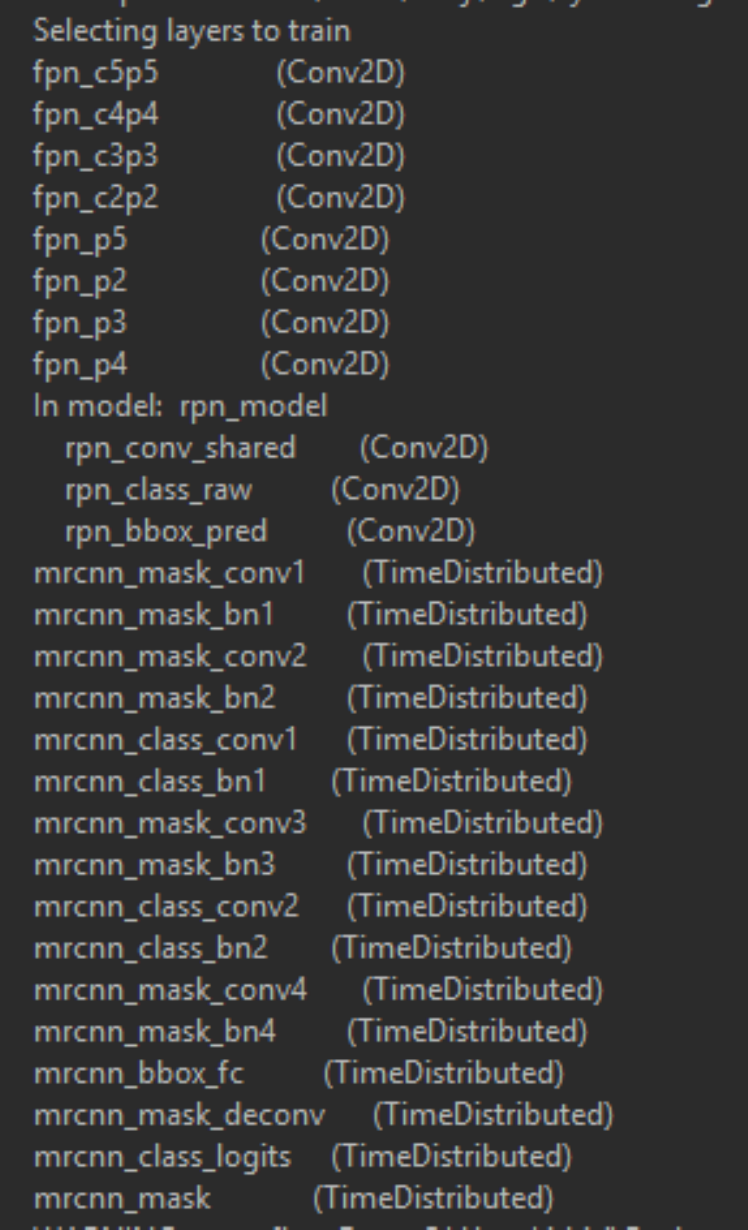}\end{center}
\caption{Layers chosen for training}
\label{fig:headlayers}
\end{figure}

One of the biggest challenges in machine learning applications in construction is the lack of enough labeled data to train the networks. To overcome this challenge, transfer learning approach can be used in which  a model developed for some task (say car detection) is reused as the starting point for a model on a second task ( hazard detection). Since low level tasks (i.e, low level feature detection) performed in any detection precess are very similar, we can re-use the weights of the most of the layers of network trained on a good dataset (eg Coco) and retrain only few out the layers with our new dataset. In this paper, Weights pre-trained on the Microsoft COCO dataset \cite{lin2014microsoft}, are used as a starting point and only the following 27 head layers are retrained: eight layers of the ResNet$-$101 backbone \cite{He2015}, three layers of the Region Proposal Network (RPN), and sixteen layers corresponding to the mask creation. The rest of the layers had their weights frozen (see Figure \ref{fig:headlayers}).
 
 Due to transfer learning, resulting accuracy is very high and training time is minimized.\\
Tensorflow \cite{tensorflow2015-whitepaper} and Keras \cite{chollet2015keras} framework are used to implement the algorithm. CUDA \cite{Nickolls:2008:SPP:1365490.1365500} and CuDNN \cite{chetlur2014cudnn} are also utilized for accelerating the computations. The whole training process took about 12 hours. The batch size is an important parameter in deep learning. A small batch size typically makes it harder for the model to converge. A large batch size can increase the efficiency by utilizing parallel computing capability and make full use of memory \cite{goyal2017accurate}. On the other hand, a large batch size would also cause a huge demand for memory. A batch size of one is used in training due to the GPU memory limitations. The model is trained for 300 epochs with 500 steps per epoch (For each epoch, the training program runs through the whole dataset once). The training requires one second per step on average. The model is trained using Keras \cite{chollet2015keras} with Tensorflow Backend on a workstation with Intel Core i7 6700K, NVIDIA GTX 1080, and 64GB RAM.

\subsection{Mapping fixations}
The objective of this step is to obtain the gaze position for every frame of the FPV video for each participant and check whether the gaze position is within any of the detected objects/area of interest (AOI). This is used to calculate how many times and for how long the subject fixates on each AOI. This data is ultimately used to determine the subject's attention distribution on the road construction site and calculate the detection likelihood score for each AOI, which is further explained in the next section.

The eye movements are captured by the wearable eye-tracking glasses (Tobii Glasses2 \cite{tobbi}). It uses infrared illuminators that illuminate the pupils and sensors that read the reflected beams to determine gaze position and direction. The eye-tracking data stream contains gaze positions of participants at every 0.01 seconds. Using the timestamps, the gaze position corresponding to each test image is extracted and subsequently checked if it is within any of the detected masks (from inference step). Since the average fixation duration in a visual search process is about 280 ms \cite{rayner2007eye}, the conditions must remain true for at least seven consecutive frames to be counted as a fixation (at 25 fps, 7 frames $\sim$ 280 ms). It is to be noted that for longer the condition might remain true for more than 280 ms or 7 frames (ie 280 is not used as a cut off). In this case, the fixation is still counted only once, and the duration of that fixation is measured by calculating the difference between the first time-stamp where the condition was met and the last time-stamp before the check fails.

\subsection{Computing eye-tracking metrics}
Once the location and duration of fixation are determined, various eye tracking metrics that define the person's viewing pattern can be computed. For demonstration purposes, only the dwell time, fixation count and on-target fixation ratio were determined, however other metrics can easily be computed as well. In the following these metrics are defined.

The fixation count defines the number of points in the visual environment where one focuses their visual attention. The dwell time for an AOI is the amount of time a person fixates on that AOI. It is calculated as the summation of all fixation durations that are detected within an AOI during the trial. Dwell time for $j_{th}$ AOI is computed as Equation \ref{equ1}, Where $DT_j$ is the dwell time for $j_{th}$ AOI, E and S are the end time and start time for $i_{th}$ fixation $(f_i)$ on $j_{th}$ AOI respectively.
\begin{equation}
DT_j  =\Sigma_{i=1}^{n} (E(f_ij )-S(f_i j))	
\label{equ1}
\end{equation}

On-target fixation ratio is the sum of fixations on AOIs (on-target fixations), divided by the total number fixations for the scene (Area of Glance or AOG). This metric indicates the amount of visual attention devoted to the hazards relative to the total attention devoted to the scene.

\section{EXPERIMENTAL SYSTEM}
To evaluate the proposed system, three participants are asked to move within a road construction site while their FPV and eye movements are recoded using a wearable eye tracking device (Tobii Glasses 2). The participants were tasked to fill out a job hazard analysis (JHA) form which mimics a standard practice of hazard recognition in construction before beginning any tasks. The objective is to map each subjects fixations with respect to the predefined objects of interest that the system is trained to detect. The system is tested using following steps:

\subsection {Test data collection}
\label{Test data collection}
\begin{figure}[H]
\begin{center}\includegraphics[width=4.5in]{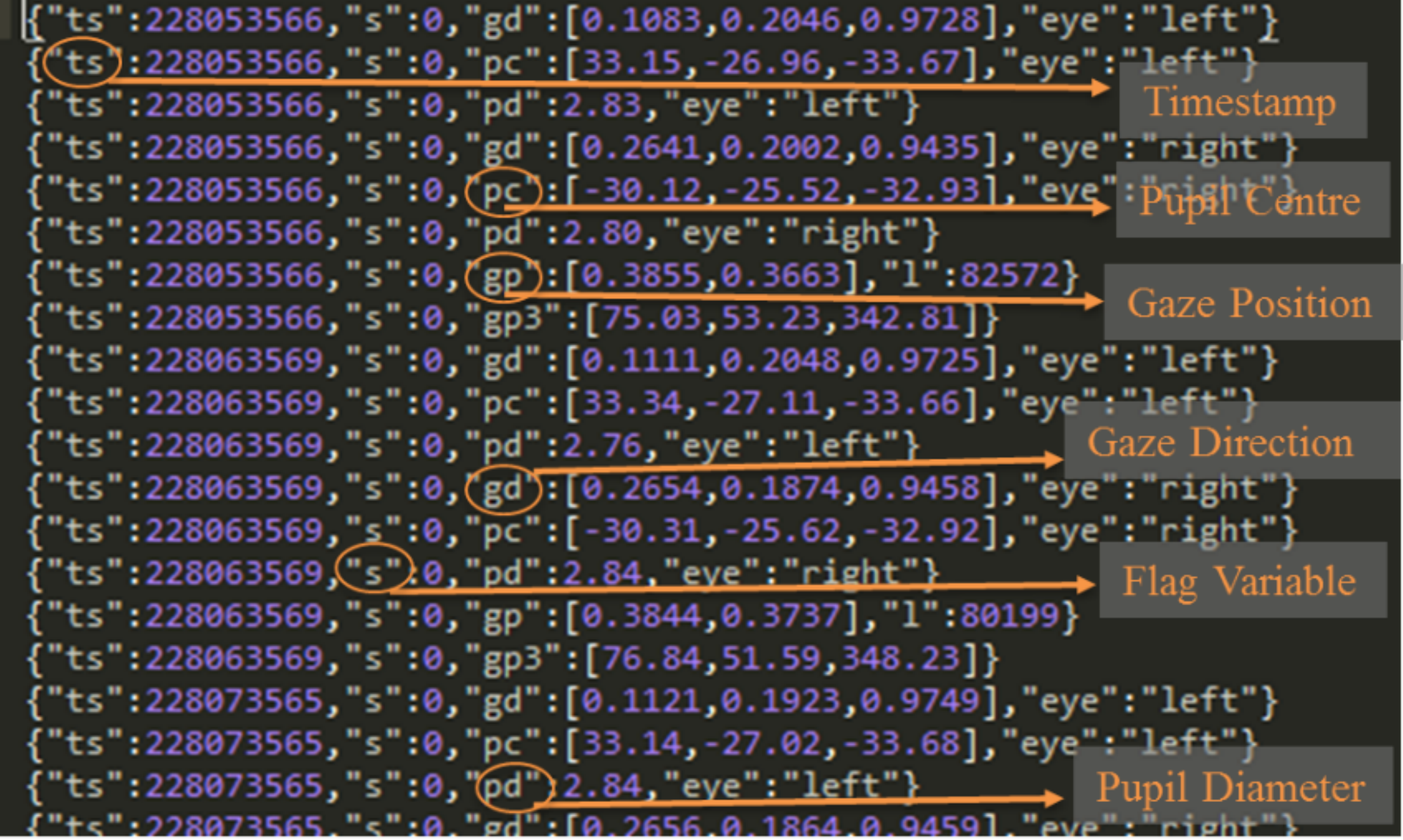}\end{center}
\caption{Excerpt from the exported raw data from Eye-tracking}
\label{fig_3}
\end{figure}

The Tobii Glasses 2 is equipped with a front mounted HD camera that records videos in first-person view (FPV). The image frames from the recorded videos are used as the test images. Table \ref{table:1} summarizes the data collected for the participants.
\begin{table}[H]
\centering
\small
\renewcommand{\arraystretch}{1}
\caption{Data collected for each participant}
\label{table:1}
\begin{tabular}{|c|c|c|c|c|}
\hline
\textbf{participant} & \textbf{Occupation} & \textbf{\begin{tabular}[c]{@{}c@{}}Calibration time\\  (seconds)\end{tabular}} & \textbf{\begin{tabular}[c]{@{}c@{}}Trial duration \\ (seconds)\end{tabular}} & \textbf{\begin{tabular}[c]{@{}c@{}}No. of collected \\ image frames\end{tabular}} \\ \hline
Participant\_1       & Student             & 11                                                                             & 28.10                                                                        & 702                                                                               \\ \hline
Participant\_2       & Student             & 9                                                                              & 26.15                                                                        & 653                                                                               \\ \hline
Participant\_3       & Student             & 12                                                                             & 25.20                                                                        & 630                                                                               \\ \hline
\end{tabular}
\end{table}
The eye movements are captured using Tobii Glasses 2 equipped with a binocular eye-tracking device with the sampling rate of 100 Hz, the visual angle of 82\degree horizontal and 52\degree vertical, and the field of view of 90\degree. The data is stored in the recording unit that is connected to the glasses via an HDMI connection. 
The raw data is exported in a JavaScript Object Notation (JSON) file (see Figure \ref{fig_3}) that contains data rows for pupil center (pc), pupil diameter (pd), gaze position in 2D (gp) and 3D (gp3), and gaze direction (gd). 

\subsection {Inference process}
\begin{figure}[H]
\begin{center}\includegraphics[width=4.5in]{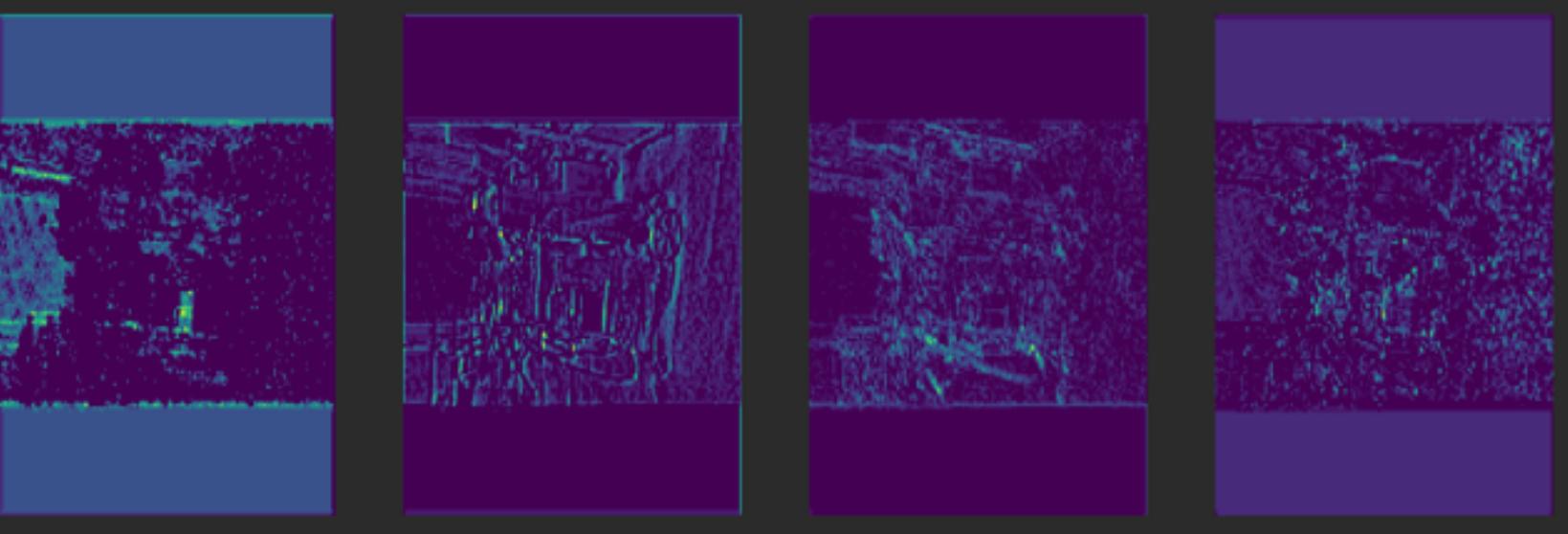}\end{center}
\caption{Output Feature Map of ResNet$-$101}
\label{fig_4}
\end{figure}

In this subsection, the inference process is explained showing the intermediate outputs for a better understanding of the whole process. The backbone here is ResNet$-$101 and this network extracts features from the input image. The output feature map is shown in Figure \ref{fig_4}.

\begin{figure}[H]
\begin{center}\includegraphics[width=6.5in]{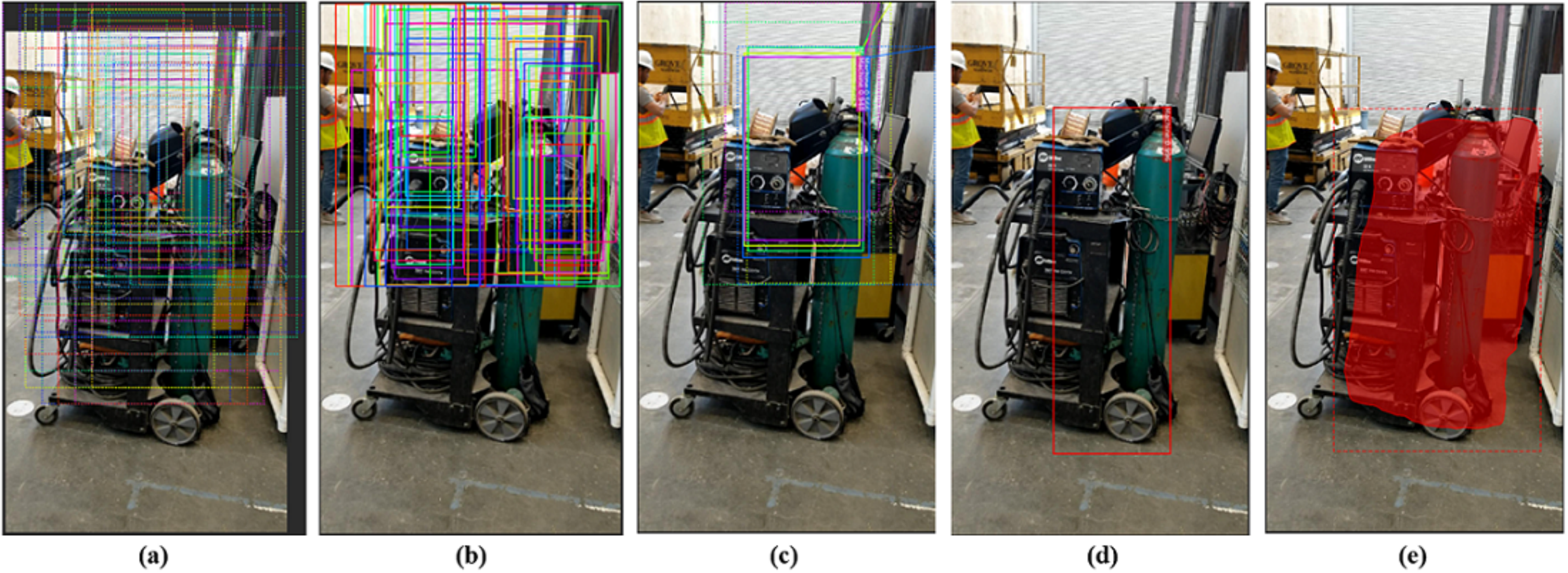}\end{center}
\caption{(a) Initial RPN prediction, unrefined anchors, (b) refined bounding boxes, (c) classified bounding boxes, (d) final bounding box, (e) mask region within the final bounding box.}
\label{fig_5}
\end{figure}

As shown in Figure \ref{fig_5}, the next step is to pass the image through the Region Proposal Network (RPN) to get the candidate regions for applying the segmentation. Figure \ref{fig_5}a shows the initial predictions from RPN. Then, the boxes are refined by removing those which extend outside image boundaries and applying non-max suppression (see Figure \ref{fig_5}b). After that, the regions are classified into one of the defined classes. As shown in Figure \ref{fig_5}c, out of the 1000 regions, only 7 regions are classified as the machine. The rest are all classified as background. Then the final region proposal is obtained (Figure \ref{fig_5}d). Finally, according to Figure \ref{fig_5}e, a mask for the region is generated and obtains the final output.

\subsection{Fixation mapping}
The fixation location is obtained for each test image from the eye tracking raw data acquired in data collection step using the timestamps. From the inference process we obtain the mask that defined the detected object location (AOI). here we check if the fixation point falls within the limits of the mask and calculate the durations using time stamps. The location and duration of fixations are used to compute various eye tracking metrics explained in previous subsections.

\section{Results}
\label{Results}
\begin{figure}[H]
\begin{center}\includegraphics[width=6.5in]{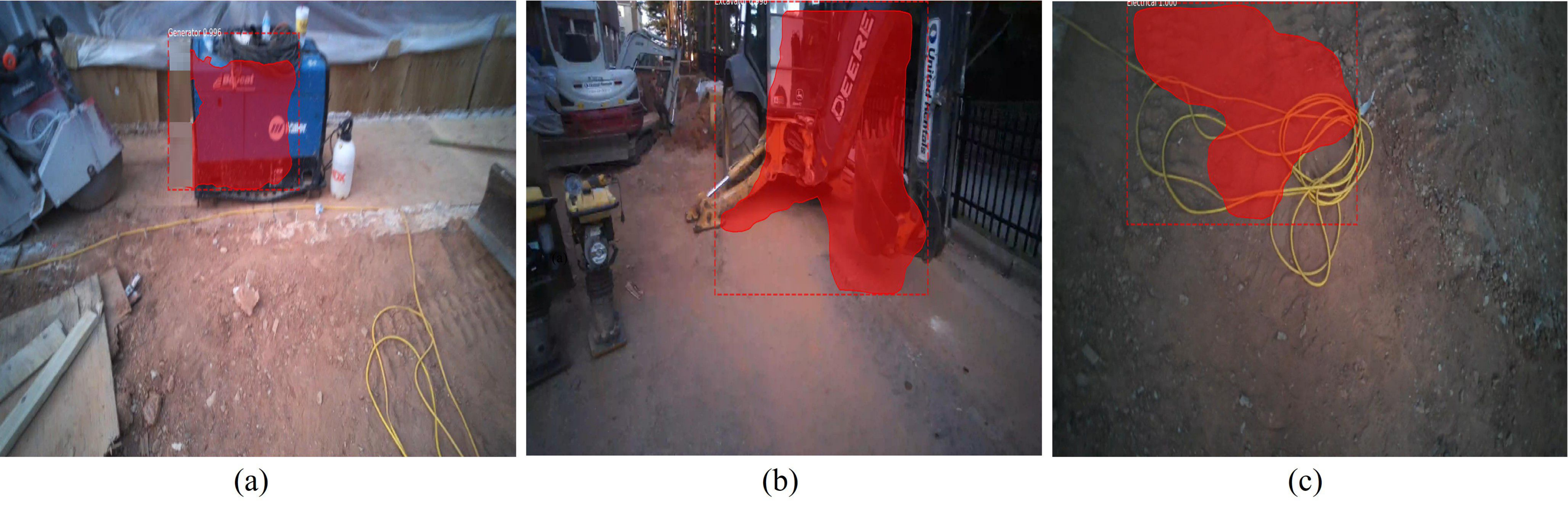}\end{center}
\caption{Inference results}
\label{fig_6}
\end{figure}

\noindent
Figure \ref{fig_6} shows detection and mask creation for three predefined hazard classes ((a) generator, (b) excavator, and (c) electric hazard) in three randomly selected frames of the test video. Various visual search metrics that define participants viewing pattern are computed as described above. Table \ref{table:2} shows the eye-tracking metrics calculated for each participant, where DT, FC, and TFR represent dwell Time, fixation count, and on-target fixation ratio respectively. 

\begin{table}[ht]
\centering
\small
\renewcommand{\arraystretch}{1}
\caption{Visual search metrics computed by the system based on fixations}
\label{table:2}
\begin{tabular}{|c|c|c|c|c|l|l|}
\hline
\textbf{participant} & \textbf{\begin{tabular}[c]{@{}c@{}}Trial duration \\ (ms)\end{tabular}} & \textbf{$DT_1$ (ms)} & \textbf{$DT_2$ (ms)} & \textbf{$DT_3$ (ms)} & FC & TFR  \\ \hline
Participant\_1       & 25410                                                                   & 5880              & 0                 & 1400              & 53 & 0.49 \\ \hline
Participant\_2       & 26150                                                                   & 3360              & 0                 & 280               & 55 & 0.23 \\ \hline
Participant\_3       & 28100                                                                   & 4200              & 0                 & 1400              & 61 & 0.32 \\ \hline
\end{tabular}
\end{table}
The results indicate that the first participant spent 25.41 seconds in the test site, and noticed H1 (the excavator) and H3 (generator). He did not pay attention to H2 (electrical hazard) H2, meaning the dwell time was zero. The participant focused on 53 distinct objects/areas relative to non-hazards, 49\% of his fixations were on AOIs (see TFC).  These metrics provide useful information about the visual search pattern of this participant. For example, the results show that out of 25410 ms of search duration, participant spent about 15000ms ($53\times280ms$ of average fixation duration) on obtaining information from the scene, that is, about 40\% of the time was spent on searching (or in saccades) during which no meaningful visual information was obtained or processed. Comparatively, a low TRC of 0.23 suggests a low level of accuracy in visual search with only 23\% of his fixations being on target. Similar feedback is generated for the other participant as well which provides more insight into their search behaviors and can be used for personalized training. It is to be noted that these results are merely for demonstration purposes. In an actual implementation of the system, several AOIs would exist and dwell times would be much higher.

\subsection{Validation}
Since \cite{DBLP:journals/corr/HeGDG17} have already validated the mask creation on AOI, this paper focuses on determining the accuracy of the system in localizing the gaze fixations on detected masks. Since all the metrics are derived from location and duration of fixations, it is important that the system correctly determines the location of the fixation in each frame. Therefore, to validate the system, 30 frames from the test videos collected from FPV of three participants are randomly selected and the fixation locations are manually obtained from the raw data for these 30 frames. These fixation locations are com[pared with the ones obtained from the system designed in this study (see table \ref{table:3}). 

The results indicate that the system accurately determines the fixation location. The minor deviation in the fixation point coordinates is due to error in synchronization of eye tracking data which is at 100 HZ and image frames extracted from 25 fps video stream, which can be easily corrected. Out of the 25 random frames that were tested, the system failed to detect the fixation on AOI (on-target fixation) for three frames. For two of the cases, it is due to the error in mask generation. This highlights one of the limitations of this method. When fixation is on the periphery of the object, the likelihood of missing the on-target fixation increases due to the error in mask generation when mask does not fully encapsulate the object. One of the misses, is because of the failure of the system to detect the object (electrical wires) due to the blurry test image. Since the camera used in this method is mounted on the eyeglasses, it is prone to sudden and fast movements (due to head movement) which results in several blurry images that limit the accuracy of the system. To overcome this we need to use eye tracking devices equipped with cameras capable of recording videos at a higher frame rate.

\begin{table}[H]
\centering
\small
\renewcommand{\arraystretch}{1.2}
\caption{Validation results: system vs ground truth}
\label{table:3}
\resizebox{\textwidth}{!}{\begin{tabular}{|c|c|c|c|c|c|}
\hline
\multirow{2}{*}{\textbf{\begin{tabular}[c]{@{}c@{}}Test \\ frames\end{tabular}}} & \multicolumn{2}{c|}{\textbf{System}}                                                                                                                                           & \multicolumn{2}{c|}{\textbf{Ground truth}}                                                                                                                                     & \multirow{2}{*}{\textbf{Remarks}} \\ \cline{2-5}
                                                                                 & \textbf{\begin{tabular}[c]{@{}c@{}}Fixation location \\ (coordinates)\end{tabular}} & \textbf{\begin{tabular}[c]{@{}c@{}}Fixation location \\ (relative to AOI)\end{tabular}} & \textbf{\begin{tabular}[c]{@{}c@{}}Fixation location \\ (coordinates)\end{tabular}} & \textbf{\begin{tabular}[c]{@{}c@{}}Fixation location \\ (relative to AOI)\end{tabular}} &                                   \\ \hline
1                                                                                & 1063,603                                                                            & Off-target                                                                               & 1063,603                                                                           & Off-target                                                                               &                                   \\ \hline
2                                                                                & 851,455                                                                             & Off-target                                                                               & 851,455                                                                             & Electrical                                                                               & Error in mask                     \\ \hline
3                                                                                & 744,533                                                                             & Off-target                                                                               & 756,545                                                                             & Off-target                                                                               &                                   \\ \hline
4                                                                                & 1343,736                                                                            & Excavator                                                                                & 1343,736                                                                            & Excavator                                                                                &                                   \\ \hline
5                                                                                & 1631,380                                                                            & Electrical                                                                              & 1631,380                                                                            & Electrical                                                                               & \multicolumn{1}{l|}{}                     \\ \hline
6                                                                                & 922,494                                                                             & Excavator                                                                                & 1,001,489                                                                           & Excavator                                                                                &                                   \\ \hline
7                                                                                & 670,688                                                                             & Excavator                                                                                & 678,692                                                                             & Excavator                                                                                &                                   \\ \hline
8                                                                                & 1433,754                                                                            & Generator                                                                                & 1433,754                                                                            & Generator                                                                                &                                   \\ \hline
9                                                                                & 1182,735                                                                            & Off-target                                                                               & 1182,735                                                                            & Off-target                                                                               &                                   \\ \hline
10                                                                               & 1393,432                                                                            & Generator                                                                                & 1393,432                                                                            & Generator                                                                                & \multicolumn{1}{l|}{}             \\ \hline
11                                                                               & 1279,738                                                                            & Generator                                                                                & 1279,738                                                                            & Generator                                                                                & \multicolumn{1}{l|}{}             \\ \hline
12                                                                               & 1062,218                                                                            & Generator                                                                                & 1062,218                                                                            & Generator                                                                                & \multicolumn{1}{l|}{}             \\ \hline
13                                                                               & 1668,580                                                                            & Off-target                                                                               & 1668,580                                                                            & Off-target                                                                               & \multicolumn{1}{l|}{}             \\ \hline
14                                                                               & 1215,324                                                                            & Excavator                                                                                & 1215,324                                                                            & Excavator                                                                                & \multicolumn{1}{l|}{}             \\ \hline
15                                                                               & 634,209                                                                             & Excavator                                                                                & 634,209                                                                             & Excavator                                                                                & \multicolumn{1}{l|}{}             \\ \hline
16                                                                               & 681,764                                                                             & Generator                                                                                & 691,766                                                                             & Generator                                                                                & \multicolumn{1}{l|}{}             \\ \hline
17                                                                               & 1144,572                                                                            & Off-target                                                                               & 1144,572                                                                            & Off-target                                                                               & \multicolumn{1}{l|}{}             \\ \hline
18                                                                               & 1256,240                                                                            & Off-target                                                                               & 1256,240                                                                            & Electrical                                                                               & Object detection failure          \\ \hline
19                                                                               & 1211,260                                                                            & Off-target                                                                               & 1211,260                                                                            & Electrical                                                                               & Error in mask                     \\ \hline
20                                                                               & 1530,731                                                                            & Generator                                                                                & 1,550,742                                                                           & Generator                                                                                &                                   \\ \hline
21                                                                               & 1439,298                                                                            & Excavator                                                                                & 1439,298                                                                            & Excavator                                                                                &                                   \\ \hline
22                                                                               & 1581,677                                                                            & Off-target                                                                               & 1,562,666                                                                           & Off-target                                                                               &                                   \\ \hline
23                                                                               & 882,682                                                                             & Excavator                                                                                & 882,682                                                                             & Excavator                                                                                &                                   \\ \hline
24                                                                               & 732,810                                                                             & Excavator                                                                                & 756,810                                                                             & Excavator                                                                                &                                   \\ \hline
25                                                                               & 683,575                                                                             & Excavator                                                                                & 672,565                                                                             & Excavator                                                                                &                                   \\ \hline

\end{tabular}}
\end{table}
\section{Conclusion}
\label{Conclusion}
The proposed system automatically maps the gaze fixations collected using a wearable eye-tracker to the real world coordinate system with respect to the predefined AOIs. The system uses transfer learning approach to train a R-CNN based model to detect various objects of interest, generate masks enclosing the detected object, and finally determine if the gaze position is within the detected AOI or not. In other words, the system detects whether and for how long a person was looking at a particular object (i.e., hazard). The data is subsequently used to calculate various eye tracking metrics that are useful to personalized training and understanding of viewing behaviors of construction workers. The system is tested on a road construction site and the validation results indicate about 88\% accuracy. The study would be of interest to practicing professionals who are interested in adopting eye-tracking in safety training that focuses on improving hazard recognition performance. The study also paves the way for future research on automating personalized safety training by combining computer vision techniques and eye-tracking technology. This study shows promising results for hazard detection using a deep learning approach.

\section{AUTHOR CONTRIBUTION STATEMENT}
The authors confirm contribution to the paper as follows: study conception and design:
Idris jeelani, Khashayar Asadi, Hariharan Ramshankar, Kevin Han, and Alex Albert; data collection: Idris Jeelani and Khashayar Asadi; analysis and interpretation of results: Idris Jeelani, Khashayar Asadi, and Kevin Han; draft manuscript preparation: Idris Jeelani, Khashayar Asadi, Hariharan Ramshankar, Kevin Han, and Alex Albert. All
authors reviewed the results and approved the final version of the manuscript.
\bibliographystyle{mytrb}
\bibliography{ref}

\end{document}